\documentclass[10pt,twocolumn]{article}

\usepackage[a4paper,margin=0.75in]{geometry}

\usepackage{cite}
\usepackage{amsmath,amssymb,amsfonts}
\usepackage{algorithmic}
\usepackage{graphicx}
\usepackage{textcomp}
\usepackage{booktabs}
\usepackage{multirow}
\usepackage{array}
\usepackage{bbm}
\usepackage{subcaption}
\usepackage{adjustbox}
\usepackage{url}
\usepackage{xcolor}
\usepackage[bottom]{footmisc}

\renewcommand{\footnoterule}{%
  \kern -3pt
  \hrule width 0.35\columnwidth
  \kern 2pt
}

\def\BibTeX{{\rm B\kern-.05em{\sc i\kern-.025em b}\kern-.08em
    T\kern-.1667em\lower.7ex\hbox{E}\kern-.125emX}}

\title{\vspace{-1.5em}
HEad and neCK TumOR (HECKTOR) 2025: Benchmark of Segmentation, Diagnosis, and Prognosis in Multimodal PET/CT
\vspace{-0.5em}}

\author{%
\parbox{0.98\textwidth}{\centering
Numan Saeed$^{1,*}$,
Salma Hassan$^{1,*}$,
Shahad Hardan$^{1,*}$,
Lishan Cai$^{2,3}$,
Xinglong Liang$^{3,4}$,
Moona Mazher$^{5}$,
Abdul Qayyum$^{6}$,
Yansong Bu$^{7,8}$,
Mengye Lyu$^{7,8}$,
Yue Lin$^{9}$,
Mingyuan Meng$^{10}$,
Chuanyi Huang$^{11}$,
Lisheng Wang$^{11}$,
Dalal Chamseddine$^{12,13}$,
Shamimeh Ahrari$^{12}$,
Beining Wu$^{14}$,
Yifei Chen$^{15}$,
Fuyou Mao$^{16}$,
Hao Zhang$^{16}$,
Baixiang Zhao$^{17,18}$,
Surajit Ray$^{17}$,
Muzi Guo$^{19}$,
Lei Xiang$^{19}$,
Jakob Dexl$^{20,21}$,
Michael Ingrisch$^{20,21}$,
Adrien Depeursinge$^{23,24}$,
Arman Rahmim$^{22}$,
Mathieu Hatt$^{25}$,
Vincent Andrearczyk$^{23,24}$,
and Mohammad Yaqub$^{1}$
}%
}
\date{} 
\begin{document}

\twocolumn[
\maketitle
\vspace{0.5em}
]

\footnotetext[1]{\scriptsize
$^{*}$Numan Saeed, Salma Hassan, and Shahad Hardan contributed equally to this work.
Corresponding author: Numan Saeed, \texttt{numan.saeed@mbzuai.ac.ae}.

$^{1}$Mohamed bin Zayed University of Artificial Intelligence (MBZUAI), Abu Dhabi, UAE.
$^{2}$Amsterdam UMC, Amsterdam, The Netherlands.
$^{3}$The Netherlands Cancer Institute, Amsterdam, The Netherlands.
$^{4}$Radboud University Medical Centre, Nijmegen, The Netherlands.
$^{5}$University College London, London, UK.
$^{6}$Imperial College London, London, UK.
$^{7}$Shenzhen Technology University, Shenzhen, China.
$^{8}$Shenzhen University, Shenzhen, China.
$^{9}$Newland Digital Technology, Fuzhou, China.
$^{10}$The University of Sydney, Sydney, Australia.
$^{11}$Shanghai Jiao Tong University, Shanghai, China.
$^{12}$University Hospital, Nantes, France.
$^{13}$Nantes Université, Centrale Nantes, CNRS, LS2N, France.
$^{14}$Hangzhou Dianzi University, Hangzhou, China.
$^{15}$Tsinghua University, Beijing, China.
$^{16}$Central South University, Changsha, China.
$^{17}$University of Glasgow, Glasgow, UK.
$^{18}$China Mobile System Integration Co., Ltd., Beijing, China.
$^{19}$Subtle Medical Inc., CA, USA.
$^{20}$University Hospital, LMU Munich, Munich, Germany.
$^{21}$Munich Center for Machine Learning, Munich, Germany.
$^{22}$BC Cancer Research Institute, Vancouver, BC, Canada.
$^{23}$HES-SO Valais-Wallis University of Applied Sciences and Arts, Sierre, Switzerland.
$^{24}$Lausanne University Hospital (CHUV), Switzerland.
$^{25}$LaTIM, INSERM, UMR 1101, Univ Brest, Brest, France.
}

\vspace{-0.1px}
\maketitle
\begin{abstract}
Head and neck cancers (HNC) represent a significant global health burden, with accurate tumor delineation being essential for effective radiotherapy planning. The complexity of the oropharyngeal anatomy, combined with the heterogeneous appearance of tumors on imaging, makes manual segmentation time-intensive and subject to inter-observer variability. Beyond segmentation, predicting long-term clinical outcomes, such as recurrence-free survival (RFS), and determining human papillomavirus (HPV) status from noninvasive imaging, remain challenging yet clinically valuable goals. The HECKTOR 2025 challenge addresses these needs by establishing a comprehensive benchmark for automated HNC analysis using multimodal PET/CT imaging and electronic health records. Building on previous editions (2020--2022), this challenge features an expanded multi-institutional dataset comprising over 1,100 patients from 10 centers worldwide. Participants were tasked with three complementary objectives: (1) segmenting primary gross tumor volumes (GTVp) and metastatic lymph nodes (GTVn), (2) predicting recurrence-free survival, and (3) classifying HPV status. The challenge attracted 35 registered teams, with 15 final submissions evaluated on a held-out test set. Top-performing algorithms achieved a mean Dice similarity coefficient of 0.75 for segmentation, a concordance index of 0.66 for survival prediction, and a balanced accuracy of 0.56 for HPV classification. This paper presents a comprehensive analysis of the submitted methodologies, evaluates their performance across different lesion characteristics, and discusses their implications for clinical translation in automated oncology workflows and decision support systems.
\end{abstract}


\section{Introduction}
\label{sec:introduction}
Head and neck cancers (HNC) are the seventh most common cancer type worldwide and are projected to reach approximately 600,000 new cases per year by 2050 \cite{sun2025global}. HNC represent a significant global health burden \cite{Deng2025Global,Hu2025Global,Hashim2019Head}, frequently arising from the mucosal surfaces \cite{2020Head} of the oral cavity, pharynx, and larynx. These malignancies are linked to a high rate of morbidity and a five-year survival rate that remains concerningly low for advanced stages \cite{Hashim2019Head}. The intricate anatomy of the head and neck (H\&N) region, characterized by a high density of vital structures and sensitive organs, makes clinical management exceptionally difficult \cite{2020Head}. Recently, the integration of multimodal imaging using positron emission tomography and computed tomography (PET/CT) has improved diagnostic accuracy and treatment selection \cite{Antoch2004Accuracyxx}. The success of these personalized treatment strategies depends heavily on the precise identification and characterization of cancerous tissues within these complex image volumes. 

Automated analysis of PET/CT images is a fundamental component of modern oncology and radiotherapy planning \cite{Andrearczyk2023Automatic,Gatidis2024Results}. In multimodal scans, the exact delineation of primary tumors (GTVp) and involved lymph nodes (GTVn) is a necessary step for calculating radiation doses \cite{Moe2021Deep}, ensuring that therapeutic beams target malignant cells while sparing healthy tissues. This process also supports the extraction of quantitative biomarkers that can inform overall clinical strategy. Beyond segmentation, interpreting these images is vital for determining disease progression and therapeutic response \cite{Hatt2016Characterization}. By extracting features from segmented volumes, clinicians can better understand the disease's biological profile, facilitating more effective monitoring and surgical or radiological navigation \cite{Postow2025Predicting,Huang2022Prediction}.

In standard practice, tumor boundaries are identified manually by specialized radiologists or radiation oncologists. In the current clinical workflow for HNC treatment, metabolic activity on PET images is often used to guide tumor contouring on the anatomical CT scan. Following this, the clinician must distinguish between primary lesions and various affected lymph nodes. When tumors are located near or infiltrate adjacent muscles and glands, experts must meticulously draw boundaries, either by reviewing axial slices or by assessing three-dimensional reconstructions \cite{Moe2021Deep}. This task is extremely time-consuming, often taking several hours for a single patient, particularly when multiple nodal involvements are present \cite{nagayasu2024enhancing}. The speed of this process is vital since treatment delays can negatively impact survival. Furthermore, the resulting contours are highly subjective and vary significantly between different medical professionals \cite{Oreiller2021Head,Andrearczyk2023Automatic}. 

For cancer prognosis and molecular characterization, the use of imaging holds great promise for supporting individualized medicine; however, it has not yet reached broad clinical implementation \cite{2016Imaging}. Predicting the risk of recurrence or identifying biological markers, such as human papillomavirus (HPV) status, is essential for tailoring treatment intensity \cite{saini2024biomarkers}. In recent research, efforts have been made to use radiomics or deep learning to predict patient outcomes from preoperative scans \cite{Bera2021Predicting,Ferro2021Prostate}. However, these methods remain difficult to translate into routine care \cite{2016Imaging}. While PET/CT is a standard diagnostic tool, extracting prognostic signals remains difficult due to the low signal-to-noise ratio in PET, the diversity of physiological uptake, and the need for robust validation across different patient groups. These obstacles underscore the necessity for advanced analytical techniques that can provide reliable prognostic information with high transparency and clinical utility \cite{Bera2021Predicting}. 

Achieving fully automated H\&N tumor analysis remains an open problem because of the wide variety of tumor sizes and locations and the subtle boundaries between cancerous and healthy tissues \cite{Illimoottil2023Recent}. Although deep learning has transformed medical image computing, its application to HNC prognosis and segmentation remains hindered by several factors \cite{Illimoottil2023Recent,Bollen2023Benefits}. This is mainly due to the difficulty in obtaining large, diverse datasets that include both high-quality segmentations and long-term clinical follow-up information \cite{Mahmood2021Artificial}. 

The Head and Neck Tumor (HECKTOR) Challenge 2025 was organized to overcome these barriers by encouraging the creation of powerful, automated tools for lesion segmentation and outcome prediction in PET/CT. Held as a satellite event during the 28th International Conference on Medical Image Computing and Computer Assisted Intervention (MICCAI 2025), this challenge sought to: (1) push the boundaries of state of the art methods for tumor and lymph node segmentation, prognosis, and diagnosis; (2) offer a high quality open access dataset to support ongoing innovation in oncological imaging; and (3) foster cooperation between computer scientists, radiologists, and oncologists.

To meet these objectives, we assembled an extensive, multi-center dataset comprising PET/CT scans from \textbf{1123} individuals across \textbf{10} different institutions. These images were collected using various scanners and protocols to represent the heterogeneity of real-world clinical settings. Expert clinicians provided ground-truth segmentation of primary tumors and lymph nodes. The dataset includes clinical variables, HPV status, and recurrence-free survival (RFS) data, with radiotherapy dose maps for a specific subset provided as additional input. Participants were asked to build fully automated models for three tasks: lesion detection and segmentation, RFS prediction, and HPV status diagnosis, thereby providing solutions that enhance the precision and effectiveness of HNC care.

Over the course of the challenge, we received 347 total submissions from 35 registered teams, representing 500+ researchers. Participation was strongly international, with contributors spanning 50+ countries worldwide. In this paper, we provide a thorough analysis of the challenge, including a detailed summary of the clinical background and related works, participation metrics, descriptions of the submitted algorithms, a comparison of performance results, and a discussion of the major findings.

\subsection{Challenges in Oncological Imaging}
Recognizing the world's increasing cancer burden \cite{globalcon}, research in artificial intelligence (AI) is increasingly focusing on the clinical value AI can deliver to aid care providers in the assessment and treatment of cancer patients. In 2025, two other MICCAI challenges were conducted on the intersection of AI and HNC: SegRap \cite{segrap}, and HANCOTHON \cite{hancothon}. SegRap \cite{segrap} is an auto-segmentation challenge for nasopharyngeal cancer that targets multi-center clinically critical volumes on CT to segment the primary tumor and lymph nodes. The dataset includes both labeled and unlabeled data to encourage robust and semi-supervised learning. The HANCOTHON challenge \cite{hancothon} explores HNC by predicting 5-year survival and 2-year recurrence risk using a single-center cohort of 763 patients and their clinical, blood, pathology, and free-text data.

\subsection{Previous HECKTOR Editions}

The HECKTOR challenge series is a multi-edition benchmark designed to advance fully automated analysis of FDG-PET/CT in oropharyngeal HNC, enabling fair comparison of methods for tumor segmentation and outcome prediction across multi-center data. It has been run across four editions (2020–2025) \cite{Oreiller2021Head, Andrearczyk2023Automatic, hecktor2022, hecktor2025}. The first edition, in 2020, focused on a single core task: automatic 3D segmentation of primary head and neck tumors from bimodal FDG-PET and CT, encouraging participants to explore effective ways to combine both modalities for robust delineation in PET-CT scans.

The HECKTOR 2021 challenge benchmarked automatic analysis of pre-treatment FDG-PET/CT \cite{Andrearczyk2023Automatic} through GTVp segmentation on PET/CT and progression-free survival prediction with and without expert contours. The dataset comprised 325 multicenter cases from six institutions, including associated clinical variables, and attracted strong community engagement. 

The HECKTOR 2022 challenge expanded the scope of HNC image analysis by requiring simultaneous segmentation of both the primary tumor volume (GTVp) and metastatic lymph nodes (GTVn), as well as RFS prediction using imaging and clinical data \cite{hecktor2022}. The dataset was scaled to 883 multi-center cases from nine institutions. 

\subsection{Automated H\&N Analysis}
\subsubsection{Segmentation}
In HNC, segmentation is significant but challenging because tumors can be irregular, close to high-FDG-uptake structures, and metastases in lymph nodes are small \cite{2020Head}. Traditional CT/PET segmentation methods rely on rule-based techniques, which are sensitive to image noise, tumor heterogeneity, and imaging acquisition protocols \cite{Oreiller2021Head}. Deep learning offers a more robust approach to segmenting GTVp and GTVn using widely adopted methods, such as U-Net \cite{unet}, nnU-Net \cite{nnunet}, and Swin UNETR \cite{swinunetr}. 

One popular method is the encoder–decoder designed U-Net \cite{unet} that extracts an abstract representation while preserving spatial detail, and skip connections that help recover fine boundaries by injecting high-resolution features into the decoding stages. Despite this, U-Net and similar convolutional designs \cite{unet} are limited in modeling global context, as long-range interactions are not explicitly captured. Motivated by this limitation, UNETR \cite{unetr} shifts 3D segmentation toward a transformer-driven formulation in which volumetric images are treated as token sequences and the prediction is in a sequence-to-sequence manner. Building on this, Swin UNETR \cite{swinunetr} integrates the Swin Transformer into the UNETR pipeline, leveraging hierarchical attention together with shifted-window mechanisms.

Combining multiple data types in medical imaging segmentation, such as PET and CT, increases the accuracy of abnormal-region localization because each modality contributes differently to the analysis \cite{deep_ctpet, fuse_petct1}. CT images mainly provide clear anatomical structure with high spatial resolution, while PET images highlight physiological activity and metabolic changes. In deep learning solutions, these modalities are fused using early, late, or hybrid fusion \cite{fusion_healthcare}. The most common fusion technique in HNC analysis is early fusion, where PET and CT are merged at the input stage \cite{hecktor2022}. While this allows direct cross-modal interaction from the start, it increases the input dimensionality and depends strongly on good registration between modalities.

\subsubsection{Prognosis}
Prognosis is a significant step that allows clinicians to estimate how the disease will progress, the likelihood of recovery, and the possibility of recurrence. Computational and clinical approaches to prognosis have evolved over time, from handcrafted features to the use of AI models to assess patient risk. Using AI for prognosis remains an active area of research, as there is a need to develop better evaluation metrics, loss functions, and methods for handling censoring \cite{censoring}.

 Classical prognosis pipelines involve segmenting the tumor region, computing engineered radiomics features from the image, and using a survival model, such as Cox proportional hazards (CoxPH) regression, to relate these features to time-to-event outcomes. Radiomics is a method for quantifying tumor phenotype using shape, first-order intensity, and texture features \cite{hnc_classical_prognosis}. However, radiomic values can vary with preprocessing choices and differences in scanners or reconstructions. After the features are extracted, the Cox PH model can be used to estimate how covariates shift the hazard rate over time via a log-linear risk score.

While classical prognostic approaches worked under the assumption of linear covariates, deep learning uncovered patterns by relaxing the restriction to linearity. One of the most popular deep learning models for survival analysis is DeepSurv \cite{deepsurv}, which replaces the linear risk score in CoxPH with a neural network and maximizes the Cox partial likelihood. However, when data can be modeled as sequences, attention-based approaches are more effective in capturing important patches and interactions across different model inputs. Thus, transformer-based methods emerged, using self-attention to model relationships and to produce a survival distribution for each patient \cite{first_transformer}. 

\subsubsection{Diagnosis}
Human papillomavirus (HPV) is a common group of viruses, and persistent infection with high-risk types can lead to several cancers, including oropharyngeal cancer \cite{hpv_cancer}. In HNC, the presence of this infection significantly affects patients' survival risk, diagnosis, and treatment \cite{prognosis_hpv}. The clinical testing of HPV status is obtained using tumor tissues from a biopsy or surgical sample \cite{hpv_cancer}. Since these procedures are invasive, AI-based imaging approaches can reduce downsides and can be repeated over time with routine scans.

Imaging-based HPV classification in oropharyngeal HNC has mainly been studied using either radiomics-driven machine learning or deep learning on routine pretreatment scans. In CT radiomics, \cite{hpv_rad} showed that features extracted from both the gross tumor and peritumoral regions can improve the prediction of HPV status. In contrast to handcrafted pipelines, \cite{hpv_3d} demonstrated that 3D convolutional neural networks can learn discriminative patterns directly from diagnostic CT to classify HPV status. Moreover, \cite{ctpet_hpv} developed PET/CT radiomics signatures from GTVp and GTVn to predict HPV association, indicating that metabolic patterns and heterogeneity on FDG-PET can be informative.

\section{Challenge Description}
\label{sec:challenge}

\begin{figure*}[!t]
\centering
\includegraphics[width=\textwidth]{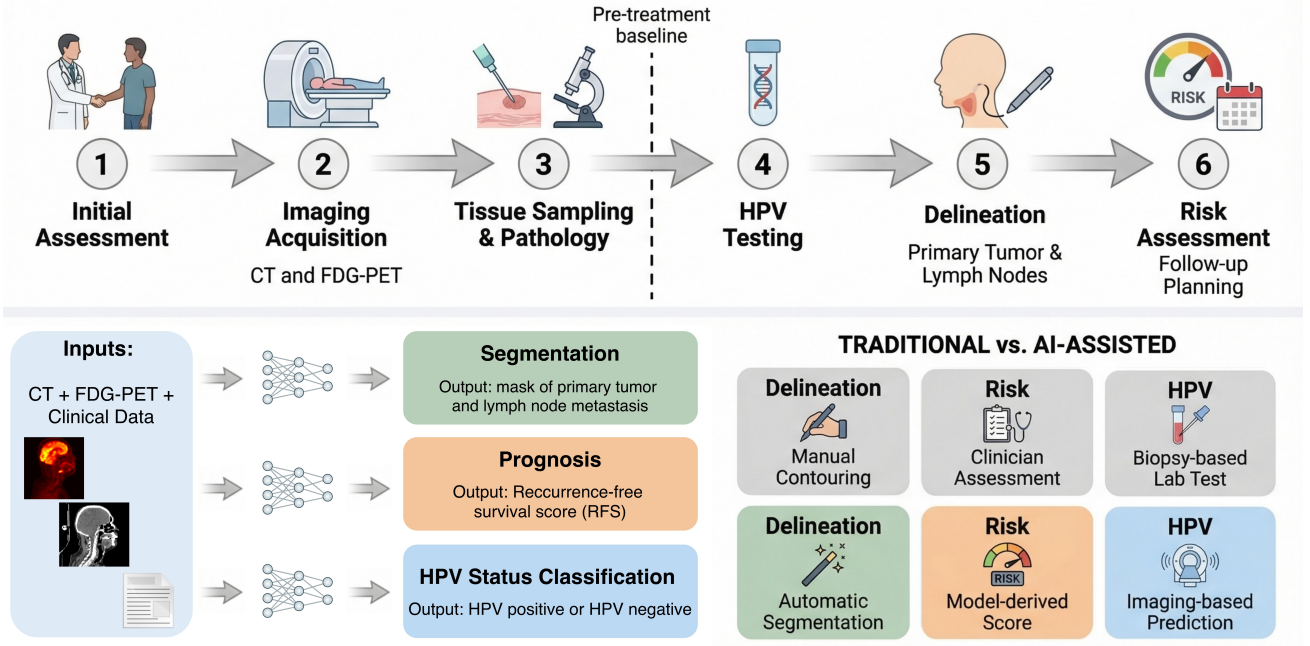}
\caption{Overview of a traditional versus AI-assisted workflow for HNC evaluation. The traditional workflow shows the clinical pathway from initial assessment, imaging, tissue sampling, HPV testing, and tumor delineation to risk assessment. AI models aid clinicians by using CT, FDG-PET, and clinical data to automate segmentation, predict prognosis, and classify HPV status non-invasively, potentially saving time and reducing costs from repetitive biopsies.}
\label{fig:qualitative}
\end{figure*}

\subsection{Challenge Setup}
The challenge consisted of three clinically significant tasks, and participants were free to enter any subset of them. Each task followed a three-phase evaluation protocol: an initial sanity-check phase, followed by validation, and, finally, a testing phase restricted to the top 15-performing teams per task on unseen data. To streamline submissions and to keep the validation and testing datasets private, ensuring fairness and reproducibility, we hosted the challenge on the Grand Challenge platform, where participants submitted their solutions as Docker containers. The challenge encompassed three main tasks: lesion segmentation and detection, prognosis, and HPV diagnosis. The task details are outlined below:

\begin{enumerate}
    \item \textbf{Segmentation Task:} The aim of this task is to delineate the primary tumor (GTVp) and metastatic lymph nodes (GTVn). The input data consist of paired 3D PET and CT scans, supplemented by selected clinical variables from electronic health records (EHR) provided in tabular form. Participants were required to produce a voxel-wise segmentation mask in which 0 denotes background, 1 denotes GTVp, and 2 denotes GTVn. In this edition, the emphasis extended beyond segmentation accuracy to also include lesion detection performance.
    \item \textbf{Prognosis Task:} This task aims to predict RFS from multimodal inputs, including PET/CT imaging and tabular EHR variables. Radiotherapy dose (RTDose) information is additionally available for a limited subset of patients. The expected output was the predicted RFS time.
    \item \textbf{HPV Diagnosis Task:} Newly introduced in this edition, this task evaluates whether imaging and non-imaging features can accurately predict HPV status in a non-invasive manner. The inputs mirror those of the segmentation task (PET/CT scans and tabular EHR variables), and the expected output was a binary HPV status prediction.

\end{enumerate}

 Participation rules discouraged the use of additional data to ensure fairness among teams. The Grand Challenge platform had a runtime environment with an NVIDIA T4 Tensor Core GPU (16 GiB VRAM), 8 vCPUs, 32 GB of memory, and a 225 GB NVMe SSD for running the container. The inference time limit per case was set to 10 minutes for segmentation and diagnosis, and 15 minutes for prognosis. Once participants submitted a Docker container, it was executed in this controlled environment against hidden validation/test cases, and the platform then ran the evaluation script to compute metrics and produce leaderboard scores.


\subsection{Data Curation and Preparation}

\subsubsection{Multi-institutional Cohort} 
\begin{figure}[t]
    \centering

    \begin{subfigure}[t]{0.20\textwidth}
        \centering
        \includegraphics[height=4.8cm]{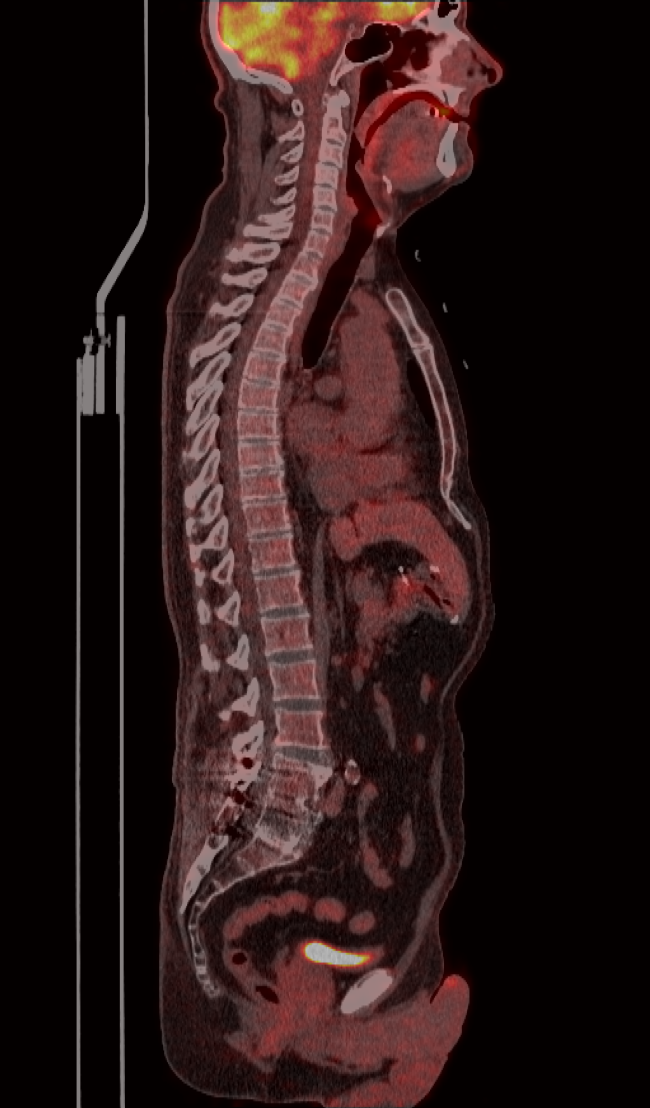}
        \caption{CHUP center}
        \label{fig:chup_center}
    \end{subfigure}
    \hfill
    \begin{subfigure}[t]{0.28\textwidth}
        \centering
        \includegraphics[height=4.8cm]{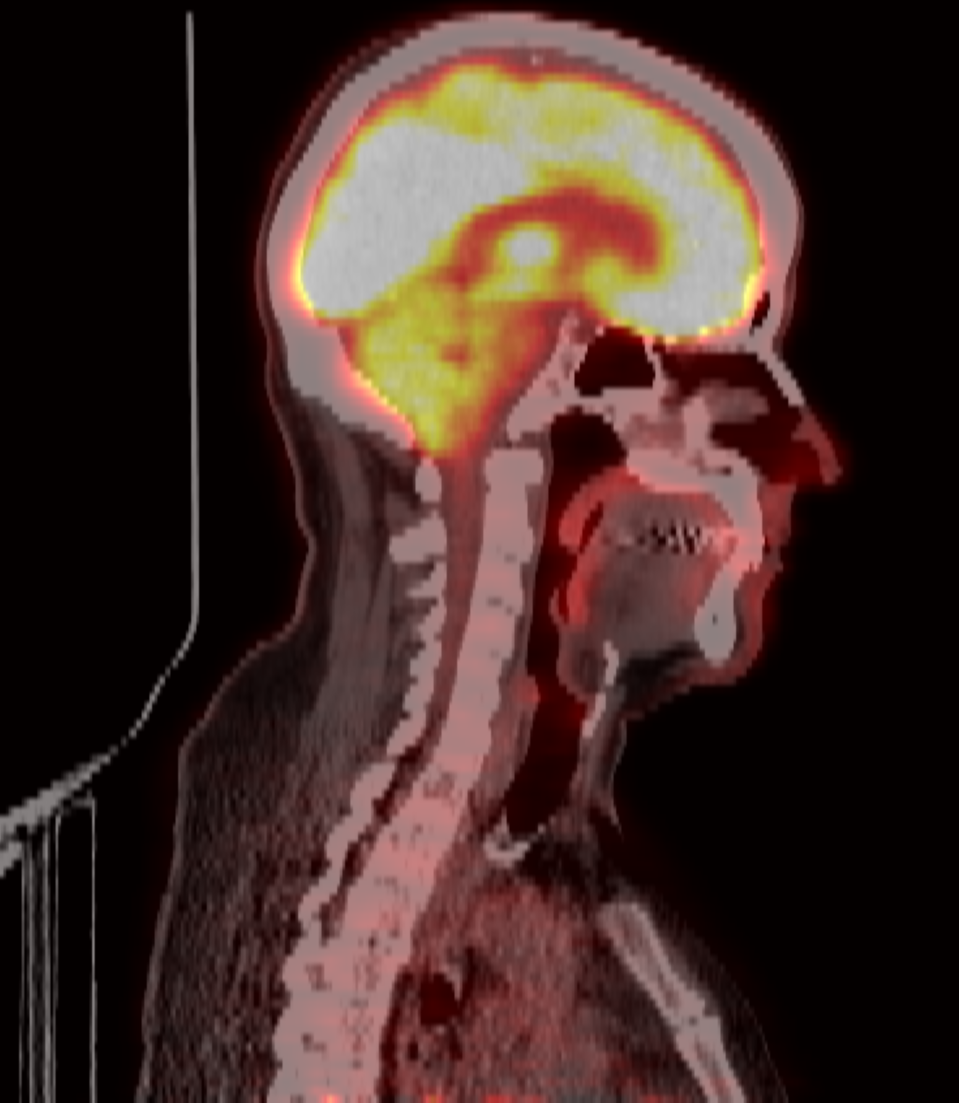}
        \caption{MDA center}
        \label{fig:mda_center}
    \end{subfigure}

    \caption{Comparison of overlayed PET/CT imaging fields of view and quality across different centers contributing to the HECKTOR dataset. (a) shows a full-body CT, while (b) are focused on the head and neck region.}
    \label{fig:ctpet}
\end{figure}

The data were collected from 1,123 patients with confirmed oropharyngeal HNC from 10 international centers across Canada, the USA, France, and Switzerland, with case counts ranging from 18 to 444 per center. FDG-PET and low-dose non-contrast CT images of the H\&N were acquired on combined PET/CT scanners across multiple sites, using a range of scanner models and manufacturers, as outlined in Table \ref{tab:centers}. Figure \ref{fig:ctpet} shows instances of overlayed CT and PET slices for a subset of centers, presenting the variety of views per center. 

Clinically standardized PET/CT protocols were used during acquisition. Across centers, patients underwent FDG PET/CT after fasting, followed by a resting uptake period. Imaging was performed using multi–bed–position acquisitions, with PET reconstruction primarily based on OSEM-family algorithms and low-dose CT for attenuation correction. In addition, RTDose was available for a limited subset of 287 patients across six centers. The RTDose provides access to delivered/planned 3D dose maps to support dose–volume metrics and RFS analyses.

The dataset includes patients aged 32 to 90 (60.7 ± 9.1) years, with approximately 70\% male and 30\% female patients. This gender imbalance reflects established epidemiological patterns. Other clinical features include HPV status, recurrence-free time, tobacco consumption, alcohol consumption, surgery, chemotherapy, radiation, performance status, and M stage.

All the DICOM scans were de-identified and converted to NIfTI format. The 1,123 cases were split into 680 for training, 50 for validation, and 393 for testing. The training data spanned 7 centers, validation was performed on a new, unseen center, and testing consisted of 5 centers, 2 of which were previously used in training and 3 were unseen. The center and data split, along with the number of cases, are summarized in Table \ref{tab:centers}.


\begin{table*}[!t]
\caption{Summary of Participating Centers, Scanner Characteristics, and Patient Counts}
\label{tab:centers}
\centering
\adjustbox{max width=\textwidth}{
\begin{tabular}{llllcc}
\toprule
\textbf{Center} & \textbf{Acronym} & \textbf{Location} & \textbf{PET/CT Scanner} & \textbf{Patients} & \textbf{Split} \\
\midrule
Centre Hospitalier Universitaire de Sherbrooke & CHUS & Canada & GeminiGXL 16, Philips & 72 & Train \\
Centre Hospitalier de L’Université de Montréal & CHUM & Canada & Discovery STE, GE Healthcare & 56 & Train \\
Centre Hospitalier Universitaire de Poitiers & CHUP & France & Biograph mCT 40 ToF, Siemens & 72 & Train \\
Hôpital Général Juif & HGJ & Canada & Discovery ST, GE Healthcare & 55 & Train \\
Hôpital Maisonneuve-Rosemont & HMR & Canada & Discovery ST, GE Healthcare & 18 & Train \\
MD Anderson Cancer Center & MDA & USA & Discovery HR, RX, ST and STE, GE Healthcare & 444 & Train/Test \\
University Hospital of Zürich & USZ & Switzerland & Discovery HR, RX, STE, LS, and 690, GE Healthcare & 101 & Train/Test \\
Centre Henri Becquerel & CHB & France & GE710, GE Healthcare & 58 & Val/Test \\
Centre Hospitalier Universitaire de Brest & CHUB & France & Philips GEMINI, Siemens Biograph & 216 & Test\\
Centre Hospitalier Universitaire de Nantes & CHUN & France & Siemens mCT 64 vision & 31 & Test \\
\midrule
\textbf{Total} & & & & \textbf{1,123} & \\
\bottomrule
\end{tabular}
}
\end{table*}

\subsubsection{Pre-processing} Data pre-processing was needed to standardize the PET data across patients and scanners to ensure comparable PET uptake and spatial alignment, so that downstream analysis is accurate. Most PET scans were provided in Standardized Uptake Value (SUV) units; however, a subset of patients had PET images in alternative units, such as becquerels per milliliter and counts, which were then converted to SUV. Additionally, no registration was required between the CT and PET images because they were acquired in the same session on an integrated PET/CT scanner. However, for the subset of patients who had RTDose scans, registration was necessary to align the RTDOSE with the reference PET/CT. Refer to \cite{hecktor2025} for more details about image acquisition and preprocessing. 


\subsubsection{Annotation Protocol} The annotation protocol focused on delineating gross tumor volume contours for GTVp and GTVn. The workflow varied across centers and by the challenge phase to obtain expert ground truth. The annotation for the old data used in the previous edition of the challenge was performed differently at each center. To ensure consistency, expert reviewers performed centralized quality control via a shared cloud environment, re-annotating contours when necessary to better reflect the true tumoral volume (often smaller than the radiotherapy planning volumes) and creating missing contours using PET/CT fusion and nodal staging information under standardized, board-defined guidelines. For the newly acquired data, the annotation workflow varied slightly. It was performed in three steps: initial automatic segmentation to generate preliminary masks of GTVp and GTVn, post-processing of the masks, and expert manual refinement of the masks using an in-house interactive PET/CT contouring tool. Refer to \cite{hecktor2025} for more details about the annotation protocol.

\subsection{Assessment Method}
The submitted algorithms were evaluated using qualitative metrics tailored to each task.
\subsubsection{Segmentation Metrics} The evaluation was designed to jointly capture delineation quality and lesion-level detection performance, since accurate lymph node analysis requires both (i) correctly localizing lesions and (ii) producing precise voxel-wise contours once detected. Participants submit fully automatic 3D masks at the original CT resolution. The submitted label map uses values \{0,1,2\} for background, GTVp, and GTVn, respectively.

The first metric used was the mean Dice similarity coefficient (DSC) on GTVp. For each patient, voxel-wise overlap between the predicted primary tumor mask $P$ and the reference mask $R$ is measured with the Dice similarity coefficient for the binary GTVp mask.
The team score for GTVp is the macro-average across all $N$ test patients:
\begin{equation}
\mathrm{DSC}^{\mathrm{GTVp}}_{\mathrm{mean}} \;=\; \frac{1}{N}\sum_{i=1}^{N}\mathrm{DSC}\!\left(P_i^{\mathrm{GTVp}}, R_i^{\mathrm{GTVp}}\right).
\end{equation}
This metric emphasizes contour accuracy of the primary tumor and penalizes both over-segmentation (extra voxels) and under-segmentation (missed voxels).

The second metric focused on nodal disease segmentation and used aggregated DSC on GTVn. Nodal disease frequently consists of multiple disconnected lesions and is therefore sensitive to how performance is aggregated across lesions and patients. To obtain a robust cohort-level summary, GTVn segmentation quality is measured with an aggregated Dice score that pools overlap statistics over the entire test set before computing Dice:
\begin{equation}
\mathrm{DSC}^{\mathrm{GTVn}}_{\mathrm{agg}} \;=\;
\frac{2\sum_{i=1}^{N}\left|P_i^{\mathrm{GTVn}}\cap R_i^{\mathrm{GTVn}}\right|}
{\sum_{i=1}^{N}\left|P_i^{\mathrm{GTVn}}\right|+\sum_{i=1}^{N}\left|R_i^{\mathrm{GTVn}}\right|}.
\end{equation}
Equivalently, this is DSC computed from the total number of overlapping nodal voxels across the full cohort. Compared with a simple per-patient mean, this aggregation reduces sensitivity to cases with very small nodal volumes and provides a stable global estimate of nodal contouring performance.

The third metric focused on nodal disease detection, with an aggregated F1-score on GTVn (lesion-level matching). Because accurate nodal modeling also depends on whether individual nodal lesions are detected at all, a lesion-level detection metric is computed in addition to voxel overlap. Let $\{p_j\}$ denote predicted nodal connected components and $\{r_k\}$ reference nodal connected components. A predicted lesion $p_j$ is considered a true positive detection if it matches at least one reference lesion $r_k$ with Intersection-over-Union (IoU) exceeding a threshold:
\begin{equation}
\mathrm{IoU}(p_j,r_k) \;=\; \frac{|p_j\cap r_k|}{|p_j\cup r_k|}
\end{equation}
A match is valid if $\mathrm{IoU}(p_j,r_k) > 0.30$. Using this matching rule over the entire test set, we accumulate the total number of $\mathrm{TP}$, $\mathrm{FP}$, and $\mathrm{FN}$, where:
\begin{itemize}
\item $\mathrm{TP}$: predicted lesions that match a reference lesion with $\mathrm{IoU}>0.30$,
\item $\mathrm{FP}$: predicted lesions with no valid match (spurious detections),
\item $\mathrm{FN}$: reference lesions with no valid matched prediction (missed detections).
\end{itemize}
We then compute cohort-level (aggregated) precision and recall to finally get the aggregated F1-score:
\begin{equation}
\begin{aligned}
\mathrm{F1}_{\mathrm{agg}}^{\mathrm{GTVn}}
&= \frac{2\,\mathrm{Precision}_{\mathrm{agg}}\,
\mathrm{Recall}_{\mathrm{agg}}}
{\mathrm{Precision}_{\mathrm{agg}}+\mathrm{Recall}_{\mathrm{agg}}} \\
&= \frac{2\,\mathrm{TP}}{2\,\mathrm{TP}+\mathrm{FP}+\mathrm{FN}} .
\end{aligned}
\end{equation}

This detection metric explicitly penalizes both missed nodal lesions (increasing $\mathrm{FN}$) and hallucinated nodal lesions (increasing $\mathrm{FP}$), independently of boundary precision.

To combine the three complementary criteria into a single final ordering, we compute a Borda count over the three metric-specific rankings: (i) $\mathrm{DSC}^{\mathrm{GTVp}}_{\mathrm{mean}}$, (ii) $\mathrm{DSC}^{\mathrm{GTVn}}_{\mathrm{agg}}$, and (iii) $\mathrm{F1}_{\mathrm{agg}}^{GTVn}$. Let $T$ be the number of teams and let $r_{t,m}\in\{1,\dots,T\}$ be the rank of team $t$ under metric $m$ (rank 1 is best). We assign Borda points $b_{t,m} \;=\; T - r_{t,m}$, and sum across the three metrics:
\begin{equation}
B_t \;=\; \sum_{m\in\{\mathrm{GTVp^{DSC}},\,\mathrm{ DSC_{agg}^{GTVn}},\,\mathrm{F1_{agg}^{GTVn}}\}} b_{t,m}.
\end{equation}
Teams are finally sorted by decreasing $B_t$, so the winner is the team with the highest total Borda score, i.e., the most consistently high-performing method across tumor segmentation, nodal segmentation, and nodal detection.


\subsubsection{Prognosis Metrics} For the prognosis tasks, the objective is to predict each patient’s RFS outcome, i.e., the time-to-event endpoint describing the time from treatment to recurrence (or last follow-up in the presence of censoring). Because RFS is a censored survival outcome, performance is assessed using Harrell's concordance index (C-index), a standard measure of discriminative ability for survival models. The C-index estimates the probability that, for a randomly chosen comparable pair of patients, the model assigns a higher predicted risk (or lower predicted survival) to the patient who experiences recurrence earlier. The teams were ranked by their final C-index scores, with the participant achieving the highest ranked first.

\subsubsection{HPV Status Diagnosis Metrics} 
For the HPV diagnosis task, the objective is to predict each patient’s HPV status as a binary label (HPV-positive vs. HPV-negative). Because class prevalence may be imbalanced in the test cohort, performance is evaluated with balanced accuracy, which weights the two classes equally by averaging sensitivity and specificity. The final ranking is based on the balanced accuracy computed on the test set; the participant with the highest balanced accuracy is ranked first. In the case of ties, the method with the higher specificity is ranked higher, prioritizing fewer false-positive HPV predictions. To compute balanced accuracy, first compute the sensitivity (true positive rate) and specificity (true negative rate), and then define balanced accuracy as the arithmetic mean of sensitivity and specificity.

\subsubsection{Ranking Scheme}
Each challenge task was evaluated and ranked independently, with no cross-task aggregation. For the tumor and lymph node segmentation task, the final leaderboard was determined using a Borda count fusion of the metric-specific rankings $\mathrm{DSC}^{\mathrm{GTVp}}_{\mathrm{mean}}$, $\mathrm{DSC}^{\mathrm{GTVn}}_{\mathrm{agg}}$, and $\mathrm{F1}_{\mathrm{agg}}^{GTVn}$. For the RFS prognosis task, teams were ranked by the C-index computed on the test cohort. For the HPV status classification task, teams were ranked by balanced accuracy on the test cohort (with specificity used as a tie-breaker).


\section{Results}
\label{sec:results}

All rankings reported in this section are task-specific and are based on the held-out final test set. Following the challenge protocol, no global ranking was computed across the three tasks. We summarize the methodological characteristics of the top-ranked submissions and present the quantitative final test results for each task separately. 

\begin{figure*}[!t]
\centering
\includegraphics[width=\textwidth]{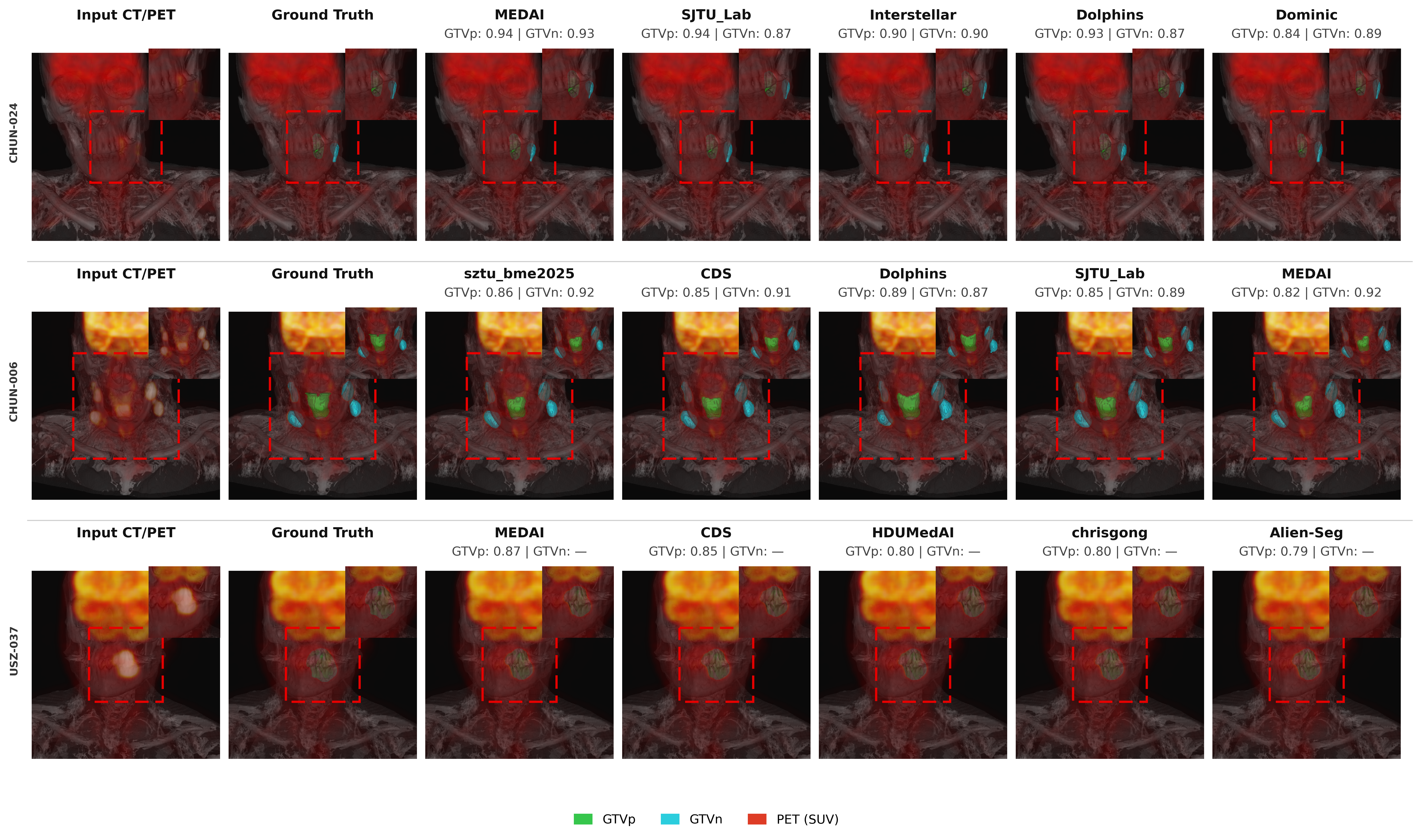}
\caption{Qualitative comparison of HECKTOR test cases with good segmentation performance. Each row corresponds to a different patient case, while columns show the input CT/PET, ground-truth annotation, and sample predictions from top-performing teams, with per-case Dice scores reported above each method. The three rows illustrate cases with a single metastatic lymph node (top), multiple metastatic lymph nodes (middle), and no metastatic lymph node (bottom), highlighting strong agreement between predictions and ground truth across different patterns of lymph node involvement.}
\label{fig:best_cases_examples}
\end{figure*}

\begin{figure*}[!t]
\centering
\includegraphics[width=\textwidth]{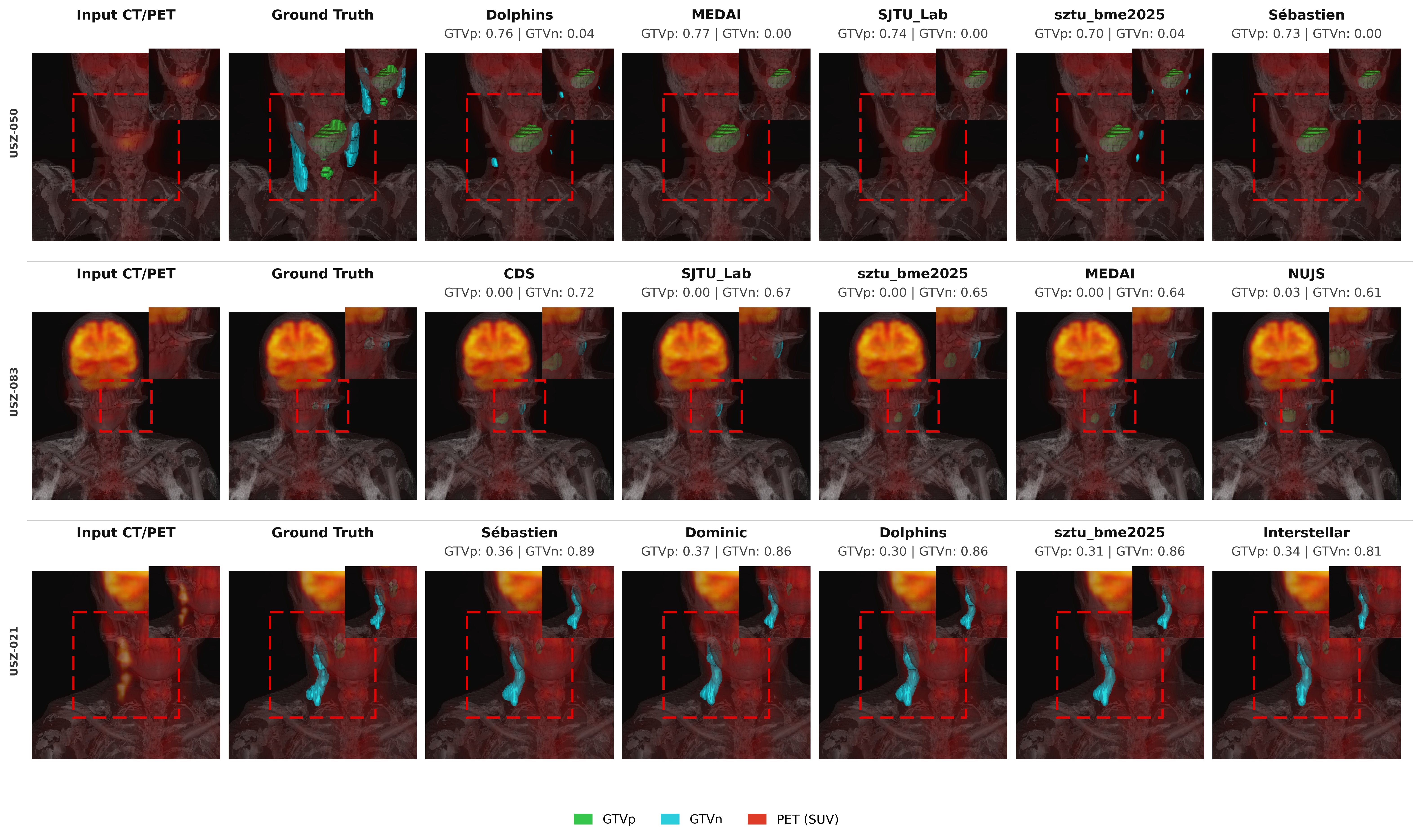}
\caption{Qualitative comparison of HECKTOR test cases with failure modes. Each row corresponds to a different patient case, while columns show the input CT/PET, ground-truth annotation (GT), and sample predictions from top-performing teams, with per-case Dice scores reported above each method. The three rows illustrate representative failure cases involving GTVn segmentation failure (top), GTVp segmentation failure (middle), and failure in both GTVp and GTVn (bottom).}
\label{fig:worst_cases_examples}
\end{figure*}




\subsection{Summary of Methodologies}
\subsubsection{Task 1: Segmentation}

\begin{table*}[!t]
\caption{Summary of the top-five Task~1 (segmentation) algorithms and their final test performance. The official rank is based on the Borda-count fusion of the three metric-specific rankings.}
\label{tab:task1_methods}
\centering
\scriptsize
\begin{adjustbox}{max width=\textwidth}
\begin{tabular}{clllllcccc}
\toprule
\textbf{Rank} & \textbf{Team} & \textbf{Backbone} & \textbf{PET/CT Fusion} & \textbf{Loss} & \textbf{Ensemble} & \textbf{GTVp DSC} & \textbf{GTVn Agg.\ DSC} & \textbf{GTVn Agg.\ F1} \\
\midrule
1 & MEDAI & STU-Net-S & Early (2-ch concat) & Dice + CE & 10-fold avg & 0.7418 & 0.7640 & 0.6472 \\
2 & Dolphins & HectoMixNet & Early (2-ch concat) & Dice + CE & 5-fold avg & \textbf{0.7567} & 0.7622 & 0.6407 \\
3 & sztu\_bme2025 & nnU-Net v2 & Early (2-ch concat) & Dice + CE & 5-fold avg & 0.7308 & \textbf{0.7641} & 0.6320 \\
4 & InterStellar & MMPL (triple-stream) & Adaptive fusion & Dice + CE & Multi-model & 0.7301 & 0.7406 & 0.6638 \\
5 & SJTU\_lab426 & nnU-Net + SegResNet & Early (4-ch concat$^\dagger$) & Dice + CE & 2-backbone avg & 0.7341 & 0.7312 & \textbf{0.7260} \\
\bottomrule
\multicolumn{9}{l}{\footnotesize $^\dagger$ Four-channel input: raw CT, squared CT, cubic-root CT, and PET.}
\end{tabular}
\end{adjustbox}
\end{table*}

\begin{table*}[!t]
\caption{Summary of the top-five Task~2 (RFS prediction) algorithms and their final test performance. The official rank is based on the concordance index.}
\label{tab:task2_methods}
\centering
\scriptsize
\begin{adjustbox}{max width=\textwidth}
\begin{tabular}{cllllllc}
\toprule
\textbf{Rank} & \textbf{Team} & \textbf{Imaging Backbone} & \textbf{Clinical Fusion} & \textbf{Auxiliary Tasks} & \textbf{Survival Model} & \textbf{Ensemble} & \textbf{C-index} \\
\midrule
1 & SIMS-LIFE & Modified 3D U-Net + 3D DenseNet & FC layers & Segmentation + HPV & Cox partial likelihood & 5-fold avg & \textbf{0.6583} \\
2 & HDUMedAI & 3D SegResNet-18 & Cross-modal fusion & Segmentation + radiomics & ICARE & -- & 0.6023 \\
2 & Alien-seg & 3D ResNet-18 & Concat + fusion MLP & None & Neural Survival predictor & 5-fold weighted avg & 0.6023 \\
4 & InterStellar & MMPL backbone & CoxPH covariates & Segmentation + HPV & CoxPH & Multi-model & 0.5873 \\
5 & MEDAI & 3D ResNet-18 + LGM & Concat + MLP & Segmentation masks & Discrete-time hazard & 5-fold avg & 0.5281 \\
\bottomrule
\end{tabular}
\end{adjustbox}
\end{table*}

\begin{table*}[!t]
\caption{Summary of the top-five Task~3 (HPV status classification) algorithms and their final test performance. The official rank is based on balanced accuracy.}
\label{tab:task3_methods}
\centering
\scriptsize
\begin{adjustbox}{max width=\textwidth}
\begin{tabular}{cllllllc}
\toprule
\textbf{Rank} & \textbf{Team} & \textbf{Imaging Backbone} & \textbf{Clinical Fusion} & \textbf{Loss} & \textbf{Class Imbalance Strategy} & \textbf{Ensemble} & \textbf{Bal.\ Acc.} \\
\midrule
1 & InterStellar & MMPL backbone (stage 2) & Radiomics + clinical indicators & Task-specific & Ensemble classifier (RF/SVM) & Multi-model & \textbf{0.5583} \\
2 & AIStat & 3D ResNet-18 & Concat + fusion MLP & CE & None reported & 5-fold avg & 0.5474 \\
3 & Villanelle & 3D ResNet-18 & Concat + FC & Weighted CE & Inverse class frequency weights & -- & 0.5380 \\
4 & MEDAI & 3D ResNet-18 + LGM & Concat + MLP & Weighted BCE & Per-sample class weights & 5-fold avg & 0.5085 \\
5 & CDS & DynUNet + HPV head & Metadata fusion & CE & Oversampling minority class & CV ensemble & 0.4908 \\
\bottomrule
\end{tabular}
\end{adjustbox}
\end{table*}

\textit{a) MEDAI ($1^{st}$ place,~\cite{1st_cai2026less})}: The authors employed STU-Net-S, the small variant of the Scalable and Transferable U-Net, which extends the standard nnU-Net architecture with residual connections to improve scalability. CT and PET volumes were resampled to isotropic $1\times1\times1$~mm$^3$ resolution and concatenated as a two-channel input. To focus the network on the relevant anatomy, a region of interest was automatically determined from PET intensities by identifying the centroid of the largest high-uptake region, and a fixed-size crop was extracted around it. CT intensities were normalized using the nnU-Net scheme, while PET values were Z-Score normalized. The model was trained with a combined Dice and cross-entropy loss to simultaneously segment GTVp and GTVn. Both five-fold and ten-fold cross-validation strategies were evaluated. The ten-fold ensemble was selected for the final submission because it improved GTVn detection on the validation leaderboard compared to the five-fold variant, an advantage attributed to increased model diversity for localizing smaller and more variable lymph node lesions. A larger model variant (STU-Net-B) was also evaluated but showed no significant performance gain despite substantially higher computational cost, supporting the authors' argument that a lightweight, well-ensembled model can outperform a heavier single architecture. 

\textit{b) Dolphins ($2^{nd}$ place, ~\cite{2nd_mazher2026hectomixnet}):} The authors proposed HectoMixNet, a 3D encoder-decoder network augmented with a bidirectional quasi separable mixing module at the bottleneck. The encoder consists of five hierarchical stages of 3D convolutional blocks with strided convolutions and batch normalization, progressively extracting multi-scale representations while preserving spatial detail through skip connections. At the bottleneck, the HectoMix module reshapes the volumetric feature maps into a sequence and applies bidirectional mixing using semiseparable state-space models, combining a causal pass that propagates information from preceding spatial positions with an anti-causal pass that aggregates context from future positions, while a learnable diagonal self-connection preserves local features. This design enables the network to integrate global spatial context without the quadratic cost of full self-attention, which is particularly important for capturing long-range dependencies between small, scattered lymph node lesions and larger primary tumors. The decoder mirrors the encoder using 3D transposed convolutions and fuses bottleneck features with multi-scale skip connections to recover fine-grained boundaries. CT and PET volumes were preprocessed with isotropic resampling, PET-intensity-based region-of-interest cropping, and modality-specific normalization. The model was trained with a combined Dice and cross-entropy loss using 5-fold cross-validation, with final predictions obtained by averaging across folds. The authors compared HectoMixNet against 3D Mamba and 3D xLSTM baselines and reported stronger lymph node segmentation scores, attributing the advantage to the bidirectional quasiseparable mixing mechanism.

\textit{c) sztu\_bme2025 ($3^{rd}$ place, ~\cite{3rd_bu2026nnunet}):} The authors applied the nnU-Net v2 framework in its default 3D full-resolution configuration without any architectural modifications or external pre-training, positioning their submission as a strong reproducible baseline. CT and PET volumes were concatenated as two input channels following the standard nnU-Net multi-modal protocol. The model was trained using five-fold cross-validation with a combined Dice and cross-entropy loss. At inference time, the standard sliding-window strategy with overlapping patches and Gaussian weighting was employed, and predictions were resampled back to the original CT geometry. Despite its simplicity, this approach ranked third overall on the final test set, demonstrating that a well-configured self-adapting framework can remain highly competitive against more complex, task-specific architectures in multi-center PET/CT segmentation.

The fourth-ranked InterStellar submission~\cite{lin2026multistage} employed a Multi-stage Multimodal Progressive Learning (MMPL) framework built on a shared triple-stream encoder with adaptive PET/CT fusion and an attention-gated decoder, where the segmentation stage served as the first of three progressive learning stages whose intermediate outputs were reused for downstream HPV classification and survival prediction.

The fifth-ranked SJTU\_lab426 submission~\cite{huang_wang_2026a} adopted a two-stage coarse-to-fine pipeline in which an nnU-Net model first localized the head-and-neck region from the full-body CT to extract a coarse ROI, after which two complementary backbones, nnU-NetResEncUNetLarge and MONAI SegResNet, were trained with five-fold cross-validation within the cropped region and their predictions ensembled. To strengthen CT-driven delineation across heterogeneous multi-center data, the authors introduced a multi-channel CT enhancement that augmented the raw CT with its squared and cubic-root intensity transforms, concatenated with PET as a four-channel input, and applied connected-component analysis to remove small isolated false positives. 

The main characteristics and final test scores for all five submissions are summarized in Table~\ref{tab:task1_methods}. Overall, the top-five Task~1 submissions exhibited substantial methodological diversity, ranging from a lightweight ensembled baseline \cite{1st_cai2026less} and a plain self-configuring framework \cite{3rd_bu2026nnunet} to architectures with novel long-range mixing modules \cite{2nd_mazher2026hectomixnet}, multi-stage progressive learning \cite{lin2026multistage}, and coarse-to-fine multi-backbone ensembles with CT enhancement \cite{huang_wang_2026a}. Despite these differences, several common design choices emerged: all methods used early fusion of PET and CT as multi-channel inputs, all employed isotropic resampling and anatomical cropping to focus on the head-and-neck region, and all relied on Dice combined with cross-entropy as the training loss. Model ensembling, whether across folds or across architectures, was a recurring strategy among the highest-ranked submissions. Notably, the plain nnU-Net v2 baseline achieved third place without any architectural modification \cite{3rd_bu2026nnunet}, suggesting that careful data handling and self-configuring training pipelines remain highly competitive in this setting.

\subsubsection{Task 2: Recurrence-Free Survival Prediction}

\textit{a) SIMS-LIFE ($1^{st}$ place,~\cite{chamseddine2026enhancing}):} The authors employed a modified version of the DeepMTS multitask framework that simultaneously performs tumor segmentation, HPV classification, and survival prediction within a unified architecture. PET and CT volumes were resampled to isotropic $2\times2\times2$~mm$^3$ resolution and cropped to $128\times128\times128$ voxels centered on the tumor using PET-intensity-based bounding boxes. The segmentation backbone, a modified 3D U-Net, produces a tumor probability map that is concatenated with the PET/CT input and fed into an HPV classification branch built on a modified 3D DenseNet, a new addition compared to the original DeepMTS. The survival branch then fuses clinical features with deep features extracted from both the segmentation encoder and the classification backbone via fully connected layers to produce a risk score, trained with the Cox partial likelihood loss. The total training objective combines segmentation (Dice plus focal), classification (focal binary cross-entropy), survival, and L2 regularization losses. To handle incomplete annotations, samples with missing segmentation or HPV labels were included during training to learn image representations but excluded from the corresponding loss terms. The authors evaluated two cross-validation strategies, a full-label five-fold scheme using only completely annotated samples and a partial-label three-fold scheme using all available data, and selected the full-label five-fold ensemble for the final submission based on its stronger test performance.

\textit{b) HDUMedAI (tied $2^{nd}$ place,~\cite{wu2026hecktor}):} The authors proposed a deep multimodal fusion network for survival prediction that combines three parallel branches: a 3D imaging encoder for CT/PET volumes, a tabular processor for structured clinical variables, and a radiomic feature branch derived from their Task~1 segmentation outputs. PET and CT volumes were first processed through HM-VNet, a hierarchical Swin-Transformer-based segmentation model with modality-specific convolutional stems and shifted-window attention, to produce anatomical priors. For the survival task, the outputs of the three branches were integrated within a cross-modal fusion module and passed into the ICARE survival prediction model \cite{rebaud2022simplicity}, which evaluates individual clinical variables and assigns binary risk weights to produce a final composite risk score.

\textit{c) Alien-seg (tied $2^{nd}$ place,~\cite{mao2026a}):} The authors developed a multimodal fusion model consisting of a 3D ResNet-18 imaging backbone that extracts a feature vector from the PET/CT input and a dedicated MLP that encodes clinical variables. The imaging and clinical feature vectors were concatenated and passed through a fusion MLP to produce a joint representation used for survival prediction. A PET-guided cropping strategy extracted a fixed region of interest around the head-and-neck area, which was subsequently resized before being fed into the 3D ResNet. The model was trained using 5-fold cross-validation, with the final prediction computed as a weighted average across the 5 folds.

The fourth-ranked InterStellar submission~\cite{lin2026multistage} reused segmentation-derived masks to guide radiomics extraction from PET/CT and combined them with the predicted HPV probability and structured clinical indicators within a Cox proportional hazards model, following the final stage of its progressive multitask framework, as described in the Task~1 summary.

The fifth-ranked MEDAI submission~\cite{1st_cai2026less} applied its multimodal survival network, a 3D ResNet-18 with a lesion-guidance module that fuses predicted segmentation masks at the earliest convolutional layer, combined with clinical features through a discrete-time hazard model, but observed a substantial generalization gap, with the C-index dropping from 0.706 on the validation leaderboard to 0.528 on the final test set.

The main characteristics and final test scores for all five submissions are summarized in Table~\ref{tab:task2_methods}. Overall, the top-five Task~2 submissions all adopted multimodal fusion strategies that combined PET/CT imaging features with structured clinical variables, confirming that imaging alone is insufficient for robust survival discrimination in this setting. The winning approach was notably the only submission to use a joint multitask objective with auxiliary segmentation and HPV classification, suggesting that shared feature learning across related clinical tasks can improve prognostic performance. The two tied second-place methods took different routes to a similar result: one leveraging transformer-based segmentation priors and radiomic features, the other relying on a simpler ResNet-MLP fusion design with ensemble averaging.

\subsubsection{Task 3: HPV Status Classification}

\textit{a) InterStellar ($1^{st}$ place,~\cite{lin2026multistage}):} For HPV classification, the authors leveraged the second stage of their Multi-stage Multimodal Progressive Learning (MMPL) framework, in which the shared triple-stream PET/CT backbone, pretrained during the segmentation stage, was progressively refined with an attention-gated decoder and an HPV prediction head. The backbone weights learned during segmentation were transferred and further optimized for the classification objective, enabling the model to reuse anatomical representations while adapting to the diagnostic task. The deep HPV score generated at this stage was combined with mask-guided radiomics features extracted from PET/CT within the predicted GTVp/GTVn contours and structured clinical indicators in an ensemble classifier to produce the final HPV probability. This progressive strategy, in which segmentation knowledge explicitly informs diagnostic prediction, enabled the method to rank first despite the task's overall difficulty.

\textit{b) AIStat ($2^{nd}$ place,~\cite{zhao2026hecktor}):} The authors applied the same unified multimodal framework used for their survival prediction model, consisting of a 3D ResNet-18 imaging branch that processes two-channel PET/CT input and a two-layer MLP that encodes clinical variables, with both feature vectors concatenated and passed through a fusion MLP to produce a joint representation. For HPV classification, a task-specific head was attached with a standard cross-entropy loss. All models were trained using stratified five-fold patient-level cross-validation with early stopping. The framework demonstrated the feasibility of a shared multimodal architecture across both prognostic and diagnostic tasks, though the final balanced accuracy of 0.5474 highlighted the difficulty of cross-center HPV classification.

\textit{c) Villanelle ($3^{rd}$ place,~\cite{guo2026hecktor}):} The authors proposed a dedicated HPV classification model with two parallel branches: an imaging branch based on a 3D ResNet-18 that extracts volumetric features from concatenated PET/CT input, and a clinical branch consisting of a two-layer fully connected network that encodes tabular variables including age, gender, tobacco and alcohol consumption, performance status, and M-stage. The imaging and clinical feature vectors were concatenated and passed through a final fully connected layer for binary prediction. To address the severe class imbalance in the training set (530 HPV-positive versus 58 HPV-negative cases), the authors adopted a weighted cross-entropy loss with class-specific weights equal to the inverse of each class's frequency. CT images were preprocessed using a mucosal windowing strategy to enhance soft-tissue contrast. Despite strong internal performance, the model exhibited a substantial generalization gap between internal validation and the organizer test set, underscoring the challenge of cross-center HPV prediction.

The fourth-ranked MEDAI submission~\cite{1st_cai2026less} reused the same multimodal framework as its survival model, combining PET/CT imaging, predicted lesion masks via the lesion-guidance module, and structured clinical variables, but replaced the survival head with a binary HPV classifier trained with a per-sample weighted cross-entropy loss, observing substantial overfitting with balanced accuracy dropping from 0.869 in cross-validation to 0.508 on the final test set.

The fifth-ranked CDS submission~\cite{dexl2026multitask} extended its segmentation model with an HPV classification head and metadata fusion, using cross-validation ensembling for the final prediction.

The main characteristics and final test scores for all five submissions are summarized in Table~\ref{tab:task3_methods}. Overall, Task~3 proved to be the most challenging of the three tasks, with the best balanced accuracy reaching only 0.558, barely above the 0.5 random baseline. All top submissions combined imaging with clinical metadata rather than relying solely on PET/CT, yet every team reported a marked generalization gap between internal or validation performance and final test-set performance. The severe class imbalance in the training data (approximately 90\% HPV-positive) and the shift to unseen centers in the test set were the primary factors limiting discriminative performance, indicating that robust non-invasive HPV classification from PET/CT remains an open problem.

\subsection{Performance Analysis}

\subsubsection{Task 1: Segmentation Performance}

The final test results for Task~1 are reported in Table~\ref{tab:task1_methods}. Under the official Borda-count ranking, MEDAI ranked first by achieving consistently high scores across all three evaluation criteria, without being the top performer on any individual metric \cite{1st_cai2026less}. The top three submissions were tightly grouped, particularly on the aggregated GTVn Dice, where the spread between the best and third-best result was only 0.0019. The metric profiles across teams were notably distinct: Dolphins achieved the highest mean GTVp Dice (0.7567) \cite{2nd_mazher2026hectomixnet}, sztu\_bme2025 obtained the highest aggregated GTVn Dice (0.7641) \cite{3rd_bu2026nnunet}, and SJTU\_lab426 achieved the highest aggregated lesion-level GTVn F1 (0.7260) \cite{huang_wang_2026a}. This pattern indicates that the final ranking was not dominated by a single metric but instead reflected a balance among primary-tumor overlap, nodal overlap, and lesion-wise nodal detection, precisely the behavior that the Borda-count fusion was designed to encourage.

A consistent finding across all submissions was that GTVn segmentation and detection were substantially more difficult than GTVp delineation. Even the best-performing method achieved a GTVn Agg.\ F1 of only 0.7260, compared to a GTVp DSC above 0.73 for most top-five entries. This gap reflects the inherent difficulty of nodal disease, in which lesions are typically smaller, more numerous, and spatially scattered across the head and neck. The lesion-level F1 metric, which explicitly penalizes both missed and hallucinated nodal lesions, proved to be the most discriminating criterion among the top submissions.

Representative favorable and failure cases are shown in Figs.~\ref{fig:best_cases_examples} and~\ref{fig:worst_cases_examples}. In the favorable cases, the leading methods exhibit strong agreement with the reference annotations across both isolated and multifocal nodal disease. In the challenging cases, errors commonly involve missed nodal lesions, under-segmentation of the primary tumor, or compound failures affecting both targets, confirming that GTVn detection remains the primary source of residual difficulty.

\subsubsection{Task 2: Survival Prediction Performance}

The final test results for Task~2 are reported in Table~\ref{tab:task2_methods}. SIMS-LIFE ranked first with a concordance index of 0.6583 \cite{chamseddine2026enhancing}, followed by HDUMedAI and Alien-seg tied at 0.6023 \cite{wu2026hecktor,mao2026a}, InterStellar at 0.5873 \cite{lin2026multistage}, and MEDAI at 0.5281 \cite{1st_cai2026less}. The gap between the winning submission and the tied second-place entries was 0.056, indicating a clearer separation at the top of the leaderboard than was observed in Task~1.

A notable observation was the generalization gap between internal cross-validation or validation-leaderboard performance and the final hidden test set. MEDAI, for example, reported a validation C-index of 0.706 but achieved only 0.528 on the test set, a drop of 0.178 \cite{1st_cai2026less}. This pattern was not unique to a single team; several submissions exhibited similar declines, suggesting that the unseen test centers introduced distribution shifts that were not captured during training. The winning SIMS-LIFE submission \cite{chamseddine2026enhancing} exhibited a comparatively smaller gap (validation of 0.583 versus test of 0.658), which may be partly attributable to its multitask training regime that provided regularization through the auxiliary segmentation and HPV classification objectives.

The overall C-index values remained moderate, with even the best submission falling below 0.66. This reflects the inherent difficulty of RFS prediction from pretreatment PET/CT, where the prognostic signal is confounded by treatment variability, incomplete clinical covariates, and heavy right-censoring in the dataset.

\subsubsection{Task 3: HPV Classification Performance}

The final test results for Task~3 are reported in Table~\ref{tab:task3_methods}. InterStellar ranked first with a balanced accuracy of 0.5583 \cite{lin2026multistage}, followed by AIStat at 0.5474 \cite{zhao2026hecktor} and Villanelle at 0.5380 \cite{guo2026hecktor}. MEDAI and CDS completed the top five with balanced accuracies of 0.5085 and 0.4908, respectively \cite{1st_cai2026less,dexl2026multitask}. The top three entries were closely grouped, with only 0.0203 separating first and third place, indicating a compressed and difficult leaderboard.

The most striking finding in Task~3 was the magnitude of the generalization gap. Villanelle reported an internal balanced accuracy of 0.9017 but achieved only 0.5380 on the test set, and MEDAI dropped from 0.869 in cross-validation to 0.508 \cite{guo2026hecktor,1st_cai2026less}. These drops are far larger than those observed in Tasks~1 and~2, and they point to two compounding factors: (1) the severe class imbalance in the training data, where approximately 90\% of cases were HPV-positive, which allows models to achieve high internal metrics by over-predicting the majority class; and (2) the center shift in the test set, where the HPV prevalence and imaging characteristics of the unseen centers may differ from the training distribution. Taken together, these results indicate that non-invasive HPV classification from PET/CT remains an open problem that will require larger and more balanced multi-center datasets, as well as more robust domain-adaptation strategies.

\subsubsection{Cross-Task Observations}

Three teams, namely MEDAI, InterStellar, and CDS, submitted to all three tasks, providing a unique opportunity to examine cross-task consistency \cite{1st_cai2026less,dexl2026multitask,lin2026multistage}. InterStellar achieved the most balanced profile, ranking $4^{th}$, $4^{th}$, and $1^{st}$ across Tasks~1, 2, and~3, respectively, which can be attributed to its progressive multitask design that explicitly transfers knowledge from segmentation to diagnosis to prognosis \cite{lin2026multistage}. MEDAI achieved the strongest segmentation performance ($1^{st}$ in Task~1) but exhibited substantial generalization drops on the downstream tasks ($5^{th}$ in Task~2, $4^{th}$ in Task~3) \cite{1st_cai2026less}, suggesting that segmentation-optimized representations do not automatically transfer to prognostic and diagnostic objectives without task-specific adaptation.

Across the three tasks, two common trends emerged. First, high final-test performance did not depend on a single architectural family: Task~1 included both highly engineered ensembles and a plain nnU-Net baseline among the top three, whereas Tasks~2 and~3 were dominated by multimodal fusion models with different backbones. Second, the strongest prognosis and HPV submissions consistently incorporated information beyond raw PET/CT intensities, either through structured clinical metadata, segmentation-derived anatomical priors, or progressive multitask learning. These results suggest that segmentation benefited primarily from strong, dense-prediction backbones and inference-time engineering, whereas prognosis and molecular classification benefited more from explicit multimodal fusion and task interdependencies.





\section{Discussion}
\label{sec:discussion}

\subsection{Analysis of Top Strategies}

The results of HECKTOR 2025 reveal several recurring design principles that contributed to strong performance across the three tasks. For segmentation, the most effective strategies shared two consistent choices: anatomical region-of-interest cropping guided by PET intensity, and the combination of Dice and cross-entropy losses to balance volumetric overlap with voxel-level classification. Of these, ensembling had the most visible impact on the leaderboard, both the first- and fifth-ranked submissions relied on it explicitly, and the third-ranked nnU-Net baseline benefited from its built-in five-fold structure. The importance of ensembling is highlighted by the fact that both the first-ranked (MEDAI \cite{1st_cai2026less}, ten-fold) and fifth-ranked (SJTU\_lab426 \cite{huang_wang_2026a}, two-backbone) submissions relied on aggregating predictions from multiple models, while the third-ranked plain nnU-Net v2 baseline benefited from its built-in five-fold ensemble. This finding is consistent with observations from previous HECKTOR editions \cite{Andrearczyk2023Automatic, hecktor2022}, where ensemble-based approaches consistently outperformed single-model submissions.

For the prognosis and HPV classification tasks, the most successful strategies moved beyond imaging-only representations by incorporating structured clinical variables and accounting for task interdependencies. The winning Task~2 submission (SIMS-LIFE) demonstrated that a multitask training objective, where segmentation and HPV classification serve as auxiliary losses, can act as an implicit regularizer that improves generalization in survival prediction \cite{chamseddine2026enhancing}. Similarly, the winning Task~3 submission (InterStellar) leveraged progressive knowledge transfer from segmentation to HPV classification, reusing anatomical representations learned during the first training stage \cite{lin2026multistage}. These findings suggest that for clinical outcome prediction in HNC, the joint modeling of related tasks offers a meaningful advantage over isolated single-task pipelines.

A notable result of this edition is that architectural complexity did not consistently translate into better performance. The plain nnU-Net v2 baseline ranked third in Task~1 despite having no custom design choices, and several architecturally elaborate submissions ranked lower on the hidden test set than on internal validation. This underscores the importance of robust preprocessing, careful data handling, and generalization-oriented training strategies over architectural novelty alone.

\subsection{Multimodal Synergy}

All top-performing submissions across the three tasks employed PET/CT fusion, confirming the complementary value of metabolic and anatomical imaging in HNC analysis. PET provides high-contrast metabolic signals that facilitate lesion detection, particularly for small lymph nodes with elevated FDG uptake, while CT offers the spatial resolution and anatomical context needed for precise boundary delineation. The dominant fusion strategy among the top segmentation submissions was early fusion through channel concatenation, where PET and CT are stacked as a multi-channel input before the first convolutional layer. This approach is simple to implement and was adopted by four of the five top-ranked Task~1 methods. The exception was InterStellar, which employed an adaptive cross-modal fusion module with separate PET and CT encoding streams, allowing the model to learn modality-specific features before merging them at intermediate pyramid levels \cite{lin2026multistage}.

For the downstream tasks (prognosis and HPV classification), the fusion challenge extended beyond imaging to include structured clinical data. All top-ranked Task~2 and Task~3 submissions incorporated clinical variables such as age, gender, tobacco and alcohol consumption, performance status, and M-stage. The most common approach was late fusion through feature concatenation, where imaging and clinical embeddings are joined before a final prediction head. However, the winning Task~2 method (SIMS-LIFE) used a deeper integration strategy in which clinical features were fused with deep features from both the segmentation and classification backbones, suggesting that richer cross-modal interactions at the feature level can improve prognostic discrimination \cite{chamseddine2026enhancing}.

\subsection{Clinical Implications}

The segmentation results achieved in Task~1, with a mean GTVp Dice of 0.76 and aggregated GTVn Dice of 0.76 for the top-ranked submission, represent a meaningful step toward clinically usable automated contouring for radiotherapy planning. These values fall within the range of inter-observer variability reported in previous studies of head and neck tumor delineation \cite{Oreiller2021Head}, suggesting that the best automated methods may be ready for integration into clinical workflows as a first-pass contouring tool, subsequently refined by radiation oncologists. The inclusion of both GTVp and GTVn targets in this edition, combined with the lesion-level detection metric, provides a more clinically complete evaluation than segmentation accuracy alone, since missed nodal lesions can directly affect treatment field design and dose coverage.

For RFS prediction, the moderate C-index values (best: 0.658) indicate that current methods provide some discriminative information beyond random ordering, but they are not yet sufficient for clinical decision-making in isolation. The prognostic models may be most useful as complementary inputs within a broader clinical decision-support framework, where imaging-derived risk scores are combined with established clinical staging systems and molecular biomarkers.

The HPV classification results present a more cautious outlook. With balanced accuracies near the 0.5 random baseline, the current generation of imaging-based HPV classifiers cannot reliably replace tissue-based testing. However, the progressive multitask approach used by InterStellar, which achieved the highest balanced accuracy by leveraging segmentation-derived features, points to a promising direction where anatomical context from tumor delineation informs molecular status prediction \cite{lin2026multistage}. Future work with larger, more balanced cohorts and explicit domain-adaptation techniques will be needed to determine whether imaging-based HPV prediction can achieve clinical utility.

\subsection{Limitations}

Several limitations of the challenge design and dataset should be considered when interpreting the results. First, the class distribution for HPV status was heavily imbalanced, making it difficult for the models to learn meaningful discriminative features for the minority class, despite the mitigation techniques implemented by several teams

Second, the incomplete availability of clinical information and imaging modalities across the cohort likely constrained model performance. Participants received only a limited set of variables, while important prognostic factors, such as smoking history and detailed treatment protocols, were not consistently available across centers. In addition, RTDose data were available for only 287 patients, restricting the usefulness of this modality for survival prediction.

Third, ground-truth segmentations were produced using heterogeneous workflows across centers, introducing potential label noise despite centralized quality control. Residual inter-annotator variability and differing contouring practices may have affected both training and evaluation, reflecting a common limitation of large multi-institutional imaging datasets.


\subsection{Future Directions}
Building on the findings and limitations of this edition, several directions merit exploration in future work and subsequent HECKTOR editions. First, addressing the class imbalance in HPV classification will require deliberate efforts to collect more HPV-negative cases across centers. Second, integrating additional data modalities, such as magnetic resonance and pathological data, offers substantial potential to improve segmentation and outcome prediction. 

Third, foundation models for medical imaging represent a promising direction for improving generalization in multi-center settings, as they could provide robust feature representations that are less sensitive to scanner and protocol variability. Finally, uncertainty quantification should be incorporated into future challenge evaluations to indicate when a model's output should be trusted and when manual review is warranted. 

\section{Conclusion}
\label{sec:conclusion}

This paper presented the HECKTOR 2025 challenge, the fourth edition in the series, organized as a satellite event at MICCAI 2025. The challenge established a comprehensive benchmark for automated head-and-neck cancer analysis using a multi-institutional, multimodal PET/CT imaging dataset comprising 1,123 patients from 10 centers. The three clinically complementary tasks are: segmentation of primary tumors and metastatic lymph nodes, prediction of RFS, and classification of HPV status.

For segmentation, the top-performing methods demonstrated that automated PET/CT-based contouring is approaching clinically relevant accuracy with employing lightweight, well-ensembled architectures and self-configuring frameworks. Lymph node detection and delineation remained the most challenging aspect, with lesion-level F1 scores substantially lower than voxel-level overlap metrics.

For predicting RFS, the best concordance index of 0.658 was achieved with a multitask learning approach that jointly optimized the three tasks. Competitive submissions confirmed that multimodal integration is essential for prognostic modeling in this setting.

For HPV status classification, a newly introduced classification in this edition, balanced accuracies remained near the random baseline, indicating the difficulty of non-invasive molecular subtyping from PET/CT, especially given the severe class imbalance and center-specific distribution shifts.

The final results, data set, evaluation code, and Docker-based submission infrastructure are publicly available at \url{https://hecktor25.grand-challenge.org} and \url{https://github.com/BioMedIA-MBZUAI/HECKTOR2025}. We anticipate that this resource will facilitate continued progress in automated HNC analysis. The challenge proceedings booklet is available at \url{https://link.springer.com/book/9783032257659}.

\section*{Acknowledgment}

The organizers thank the Centre Hospitalier Universitaire de Sherbrooke, Centre Hospitalier de l'Universit\'{e} de Montr\'{e}al, Centre Hospitalier Universitaire de Poitiers, H\^{o}pital G\'{e}n\'{e}ral Juif, H\^{o}pital Maisonneuve-Rosemont, MD Anderson Cancer Center, University Hospital of Z\"{u}rich, Centre Henri Becquerel, Centre Hospitalier Universitaire de Brest, Centre Hospitalier Universitaire de Nantes, and Sheikh Shakhbout Medical Center for providing imaging data and clinical annotations. We also thank the Grand Challenge platform and all participating teams. This work was supported in part by Mohamed bin Zayed University of Artificial Intelligence and the Society of Nuclear Medicine and Molecular Imaging.

\noindent \textbf{Conflict of Interest:} None of the authors has a conflict of interest to disclose.

\bibliographystyle{IEEEtran}
\bibliography{references}






\end{document}